\documentclass{article}
\usepackage{ijcai19}
\usepackage{amssymb}

\usepackage{times}
\usepackage{soul}
\usepackage{url}
\usepackage[hidelinks]{hyperref}
\usepackage[utf8]{inputenc}
\usepackage[small]{caption}
\usepackage{graphicx}
\usepackage{amsmath}
\usepackage{booktabs}
\usepackage{algorithm}
\usepackage{algorithmic}
\title{Requisite Variety in \textit{Ethical} Utility Functions for AI Value Alignment}

\author{
Nadisha-Marie Aliman $^1$
\and
Leon Kester $^2$
\affiliations
$^1$ Utrecht University, Utrecht, Netherlands\\
$^2$ TNO Netherlands, The Hague, Netherlands
\emails
\
nadishamarie.aliman@gmail.com
}

\begin{document}

\maketitle

\begin{abstract}
Being a complex subject of major importance in AI Safety research, value alignment has been studied from various perspectives in the last years. However, no final consensus on the design of ethical utility functions facilitating AI value alignment has been achieved yet. Given the urgency to identify systematic solutions, we postulate that it might be useful to start with the simple fact that for the utility function of an AI not to violate human ethical intuitions, it trivially has to be \textit{a model} of these intuitions and reflect their \textit{variety} -- whereby the most accurate models pertaining to human entities being biological organisms equipped with a brain constructing concepts like moral judgements, are \textit{scientific} models. Thus, in order to better assess the variety of human morality, we perform a transdisciplinary analysis applying a security mindset to the issue and summarizing variety-relevant background knowledge from neuroscience and psychology. We complement this information by linking it to \textit{augmented utilitarianism} as a suitable ethical framework. Based on that, we propose first practical guidelines for the design of approximate \textit{ethical goal functions} that might better capture the variety of human moral judgements. Finally, we conclude and address future possible challenges.
\end{abstract}

\section{Introduction}
AI value alignment, the attempt to implement systems adhering to human ethical values has been recognized as highly relevant subtask in AI Safety at an international level and studied by multiple AI and AI Safety researchers across diverse research subareas~\cite{hadfield2016cooperative,soares2017agent,yudkowsky2016ai} (a review is provided in~\cite{taylor2016alignment}). Moreover, the need to investigate value alignment has been included in the Asilomar AI Principles~\shortcite{asilomar2018principles} with a worldwide support of researchers from the field. While value alignment has often been tackled using reinforcement learning~\cite{abel2016reinforcement} (and also reward modeling~\cite{leike2018scalable}) or inverse reinforcement learning~\cite{abbeel2004apprenticeship} methods, we focus on the approach to explicitly formulate cardinal ethical utility functions crafted by (a representation of) society and assisted by science and technology which has been termed \textit{ethical goal functions}~\cite{Delphi,werkhoven2018telling}. In order to be able to formulate utility functions that do not violate the ethical intuitions of most entities in a society, these ethical goal functions will have to be a model of human ethical intuitions. This simple but important insight can be derived from the good regulator theorem in cybernetics~\cite{conant1970every} stating that \textit{``every good regulator of a system must be a model of that system"}. We believe that instead of learning models of human intuitions in their apparent complexity and ambiguity, AI Safety research could also make use of the already available scientific knowledge on the nature of human moral judgements and ethical conceptions as made available e.g.\ by neuroscience and psychology. The human brain did not evolve to facilitate rational decision-making or the experience of emotions, but instead to fulfill the core task of allostasis (anticipating the needs of the body in an environment before they arise in order to ensure growth, survival and reproduction)~\cite{barrett2017emotions,kleckner2017evidence}. Thereby, psychological functions such as cognition, emotion or moral judgements are closely linked to the predictive regulation of physiological needs of the body~\cite{kleckner2017evidence} making it indispensable to consider the embodied nature of morality when aspiring to model it for AI value alignment.\par For the purpose of facilitating the injection of requisite knowledge reflecting the variety of human morality in ethical goal functions, Section~\ref{var} provides information on the following variety-relevant aspects: 1) the essential role of affect and emotion in moral judgements from a modern constructionist neuroscience and cognitive science perspective followed by 2) dyadic morality as a recent psychological theory on the nature of cognitive templates for moral judgements. In Section~\ref{aug}, we propose first guidelines on how to approximately formulate ethical goal functions using a recently proposed non-normative socio-technological ethical framework grounded in science called \textit{augmented utilitarianism}~\cite{aliman2019augmented} that might be useful to better incorporate the requisite variety of human ethical intuitions (especially in comparison to classical utilitarianism). Thereafter, we propose how to possibly validate these functions within a socio-technological feedback-loop~\cite{Delphi}. Finally, in Section~\ref{conc}, we conclude and specify open challenges providing incentives for future work.

\section{Variety in Embodied Morality}
\label{var}
 \begin{figure}
	\centering
	\includegraphics[height=0.3\textheight]{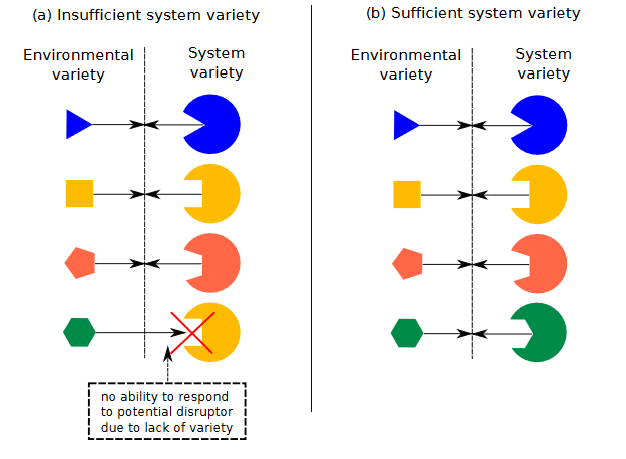}
	\caption[9pt]{Intuitive illustration for the Law of Requisite Variety. Taken from~\cite{norman2018special}. }
	\label{pic}
\end{figure}

While value alignment is often seen as a safety problem, it is possible to interpret and reformulate it as a related security problem which might offer a helpful different perspective on the subject emphasizing the need to capture the variety of embodied morality. One possible way to look at AI value alignment is to consider it as being an attempt to achieve advanced AI systems exhibiting adversarial robustness against malicious adversaries attempting to lead the system to action(s) or output(s) that are perceived as violating human ethical intuitions. From an abstract point of view, one could distinguish different means by which an adversary might achieve successful attacks: e.g. 1) by fooling the AI at the perception-level (in analogy to classical adversarial examples~\cite{goodfellow2018defense}, this variant has been denoted \textit{ethical adversarial examples}~\cite{aliman2019augmented}) which could lead to an unethical behavior even if the utility function would have been aligned with human ethical intuitions or 2) simply by disclosing dangerous (certainly unintended from the designer) unethical implications encoded in its utility function by targeting specific mappings from perception to output or action (this could be understood as ethical adversarial examples on the utility function itself). While the existence of point 1) yields one more argument for the importance of research on adversarial robustness at the perception-level for AI Safety reasons~\cite{goodf} and a sophisticated combination of 1) and 2) might be thinkable, our exemplification focuses on adversarial attacks of the type 2). \par
One could consider the explicitly formulated utility function $ U $ as representing a separate model\footnote{a conceptually similar separation of objective function model and optimizing agent has been recently performed for reward modeling~\cite{leike2018scalable}} that given a sample, outputs a value determining the perceived ethical desirability of that sample which should ideally be in line with the society that crafted this utility function. The attacker which has at his disposal the knowledge on human ethical intuitions, can attempt targeted misclassifications at the level of a single sample or at the level of an ordering of multiple samples whereby the ground-truth are the ethical intuitions of most people in a society. The Law of Requisite Variety from cybernetics~\cite{ashby1961introduction} states that \textit{``only variety can destroy variety"}, with other words in order to cope with a certain variety
of problems or environmental variety, a system needs to exhibit a suitable and sufficient
variety of responses. Figure~\ref{pic} offers an intuitive explanation of this law. Transferring it to the mentioned utility function $ U $, it is for instance conceivable that if $ U $ does not encode affective information that might lead to a difference in ethical evaluations, an attacker can easily craft a sample which $ U $ might misclassify as ethical or unethical or cause $ U $ to generate a total ordering of samples that might appear unethical from the perspective of most people. Given that $ U $ does not have an influence on the variety of human morality, the only way to respond to the disturbances of the attacker and reduce the variety of possible undesirable outcomes, is by increasing the own variety -- which can be achieved by encoding more relevant knowledge.   

\subsection{Role of Emotion and Affect in Morality }
One fundamental and persistent misconception about human biology (which does not only affect the understanding of the nature of moral judgements) is the assumption that the brain incorporates a layered architecture in which a battle between emotion and cognition is given through the very anatomy of the \textit{``triune brain"}~\cite{maclean1990triune} exhibiting three hierarchical layers: a reptilian brain on top of which an emotional animalistic paleomammalian limbic system is located and a final rational neomammalian cognition layer implemented in the neocortex. This flawed view is not in accordance with neuroscientific evidence and understanding~\cite{barrett2017emotions,miller2018happily}. In fact, the assumed reactive and animalistic limbic regions in the brain are predictive (e.g.\ they send top-down predictions to more granular cortical regions), control the body as well as attention mechanisms while being the source of the brain's internal model of the body~\cite{barrett2015interoceptive,barrett2017theory}.\par  Emotion and cognition do not represent a dichotomy leading to a conflict in moral judgements~\cite{helion2015beyond}. Instead, the
distinction between the experience of an instance of a concept as belonging to the category of emotions versus the category of cognition is grounded in the focus of attention of the brain~\cite{barrett2015conceptual} whereby \textit{``the experience of cognition
	occurs when the brain foregrounds mental contents
	and processes" } and \textit{``the experience of emotion
	occurs when, in relation to the current situation, the
	brain foregrounds bodily changes"} ~\cite{hoe}. The mental
phenomenon of actively dynamically simulating different alternative scenarios (including anticipatory emotions) has also been termed conceptual consumption~\cite{gilbert2007prospection} and plays a role in decision-making and moral reasoning. While emotions are discrete constructions of the human brain, core affect allows a low-dimensional experience of interoceptive sensations (sensory array from within the body) and is a continuous property of conciousness with the dimensions of valence (pleasantness/unpleasantness) and arousal (activation/deactivation)~\cite{kleckner2017evidence}.  It has been argued that core affect provides a basis for moral judgements in which different events are qualitatively compared to each other~\cite{cabanac2002emotion}. Like other constructed mental states, moral judgements involve domain-general brain processes which simply put combine 1) the interoceptive sensory array, 2) the exteroceptive sensory inputs from the environment and 3) past experience/ knowledge for a goal-oriented situated conceptualization (as tool for allostasis)~\cite{oosterwijk2012states}.  From these key constituents of mental constructions one can extract the following:  concepts (including morality) are\textit{ perceiver-dependent} and \textit{time-dependent}. Thereby, affect, (but not emotion~\cite{cameron2015constructionist}) is a necessary ingredient of every moral judgement. More fundamentally, \textit{``the human brain is anatomically structured so that no decision or action can be free of interoception and affect"}~\cite{barrett2017emotions} -- this includes any type of thoughts that seem to correspond to the folk terms of ``rational"  and ``cold". Therefore, a utility function without affect-related parameters might not exhibit a sufficient variety and might lead to the violation of human ethical intuitions.\par  Morality cannot be separated from a model of the body, since the brain constructs the human perception of reality based on what seems of importance to the brain for the purpose of allostasis which is inherently strongly linked to interoception~\cite{barrett2017emotions}. Interestingly, even the imagination of future not yet experienced events is facilitated through situated recombinations of sensory-motor and affective nature in a similar way as the simulation of actually experienced events~\cite{addis2018episodic}. To sum up, there is no battle between emotion and cognition in moral judgements. Moreover, there is also no specific moral faculty in the brain, since moral judgements are based on domain-general processes within which affect is always involved to a certain degree. One could obtain insufficient variety in dealing with an adversary crafting ethical adversarial examples on a utility model $ U $ if one ignores affective parameters. Further crucial parameters for ethical utility functions could be e.g.\ of cultural, social and socio-geographical nature. \par

\subsection{Variety through ``Dyadicness"}

The psychological theory of dyadic morality~\cite{schein2018theory} posits that moral judgements are based on a fuzzy cognitive template and related to the perception of an intentional agent ($ iA $) causing damage ($ d $) to a vulnerable patient ($ vP $) denoted $ iA\xrightarrow{\text{d}}vP $. More precisely, the theory postulates that the perceived immorality of an act is related to the following three elements: norm violations, negative affect and importantly perceived harm. According to a study, the reaction times in describing an act as immoral predict the reaction times in categorizing the same act as harmful~\cite{schein2015unifying}. The combination of these basic constituents is suggested to lead to the emergence of a rich diversity of moral judgements~\cite{gray2017think}. \textit{Dyadicness} is understood as a continuum predicting the condemnation of moral acts. The more a human entity perceives an intentional agent inflicting damage to a vulnerable patient, the more immoral this human perceives the act. As stated by Schein and Gray, the dyadic harm-based cognitive template \textit{``is  rooted  in  innate  and evolved processes of the human mind; it is also shaped by cultural learning, therefore allowing cultural pluralism"}. Importantly, the nature of this cognitive template reveals that moral judgements besides being perceiver-dependent, might vary across diverse parameters such as especially e.g.\ in relation to the perception of agent, act and patient in the outcome of the action. Further, the theory also foresees a possible time-dependency of moral judgements by introducing the concept of a \textit{dyadic  loop}, a feedback cycle resulting in an iterative polarization of moral judgements through social discussion modulating the perception of harm as time goes by. Overall, moral judgements are understood as constructions in the same way visual perception, cognition or emotion are constructed by the human mind. Similarly to the existence of variability in visual perception, variability in morality is the norm which often leads to moral conflicts~\cite{schein2016visual}. However, the understanding that humans share the same harm-based cognitive template for morality has been described as reflecting \textit{``cognitive unity in the variety of perceived harm"}~\cite{schein2018theory}.\par 
Analyzing the cognitive template of dyadic morality, one can deduce that human moral judgements do not only consider the outcome of an action as prioritized by consequentialist frameworks like classical utilitarianism, nor do they only consider the state of the agent which is in the focus of virtue ethics. Furthermore, as opposed to deontological ethics, the focus is not only on the nature of the performed action. The main implications for the design of utility functions that should ideally be aligned with human ethical values, is that they might need to encode information on agent, action, patient as well as on the perceivers -- especially with regard to the cultural background. This observation is fundamental as it indicates that one might have to depart from classical utilitarian utility functions $ U(s') $ which are formulated as total orders at the abstraction level of outcomes i.e.\ states (of affairs) $ s'  $. In line with this insight, is the context-sensitive and perceiver-dependent type of utility functions considering agent, action and outcome which has been recently proposed within a novel ethical framework denoted \textit{augmented utilitarianism}~\cite{aliman2019augmented} (abbreviated with AU in the following). Reconsidering the dyadic morality template $ iA\xrightarrow{\text{d}}vP $, it seems that in order to better capture the variety of human morality, utility functions -- now transferring it to the perspective of AI systems -- would need to be at least formulated at the abstraction level of a \textit{perceiver-dependent} evaluation of a transition $ s\xrightarrow{\text{a}}s'$ leading from a state $ s  $ to a state $ s' $ via an action $ a $. We encode the required novel type of utility function with $ U_x(s,a,s') $ with $ x $ denoting a specific perceiver. This formulation could enable an AI system implemented as utility maximizer to jointly consider parameters specified by a perceiver which are related to its perception of agent, the action and the consequences of this action on a patient. Since the need to consider time-dependency has been formulated, one would consequently also require to add the time dimension to the arguments of the utility function leading to $ U_x((s,a,s'),t) $.

\section{Approximating Ethical Goal Functions}
\label{aug}

While the psychological theory of dyadic morality was useful to estimate the abstraction level at which one would at least have to specify utility functions, the closer analysis on the nature of the construction of mental states performed in Section~\ref{var}, abstractly provides a superset of primitive relevant parameters that might be critical elements of every moral judgement (being a mental state). Given a perceiver $ x $, the components of this set are the following subsets: 1) parameters encoding the interoceptive sensory array $ B_x $ (from within the body) which are accessible to the human consciousness via the low-dimensional core affect, 2) the exteroceptive sensory array $ E_x $ encoding information from the environment and 3) the prior experience $ P_x $ encoding memories. Moreover, these set of parameters obviously vary in time. However, to simplify, it has been suggested within the mentioned AU framework, that ethical goal functions will have to be updated regularly (leading to a so-called socio-technological feedback-loop~\cite{Delphi}) in the same way as votes take place at regular intervals in a democracy. One could similarly assume that this regular update will be sufficient to reflect a relevant change in moral opinion and perception. 

\subsection{Injecting Requisite Variety in Utility}
For simplicity, we assume that the set of parameters $ B_x $, $ P_x $ and $ E_x $ are invariant during the utility assignment process in which a perceiver $ x $ has to specify the ethical desirability of a transition $ s\xrightarrow{\text{a}}s'$ by mapping it to a cardinal value $U_x(s,a,s')$ obtained by applying a not-nearer defined type of scientifically determined transformation $ v_x $ (chosen by $ x $) on the mental state of $ x $. This results in the following naive and simplified mapping however adequately reflecting the property of \textit{mental-state-dependency} formulated in the AU framework (the required dependency of ethical utility functions on parameters of the own mental state function $ m_x $ in order to avoid perverse instantiation scenarios~\cite{aliman2019augmented}):
\begin{align}
U_x(s,a,s')= v_x(m_x((s,a,s'), B_x, P_x, E_x))
\end{align}%

Conversely, the utility function of classical utilitarianism is only defined at the impersonal and context-independent abstraction level of $ U(s') $ which has been argued to lead to both \textit{perverse} instantiation problem but also to the \textit{repugnant} conclusion and related impossibility theorems in population ethics for consequentialist frameworks which do not apply to mental-state-dependent utility functions~\cite{aliman2019augmented}. The idea to restrict human ethical utility functions to the considerations of outcomes of actions alone -- ignoring affective parameters of the own current self -- as practiced in classical utilitarianism while later referring to the resulting total orders with emotionally connoted adjectives such as ``repugnant" or ``perverse" has been termed the \textit{perspectival fallacy of utility assignment}~\cite{Delphi}. The use of consequentialist utility functions affected by the impossibility theorems of Arrhenius~\shortcite{arrhenius2000impossibility} has been justifiably identified by Eckersley~\shortcite{eckersley2018impossibility} as a safety risk if used in AI systems without more ado. It seems that the isolated consideration of outcomes of actions (for consequentialism) or actions (for deontological ethics) or the involved agents (for virtue ethics) does not represent a good model of human ethical intuitions. It is conceivable, that if a utility model $ U $ is defined as utility function $ U(s') $, the model cannot possibly exhibit a sufficient variety and might more likely violate human ethical intuitions  than if it would be implemented as a context-sensitive utility function $ U_x(s,a,s') $. (Beyond that, it  has been argued that consequentialism implies the rejection of \textit{``dispositions and emotions, such as regret, disappointment, guilt and resentment"} from ``rational" deliberation ~\cite{verbeek2001consequentialism} and should i.a.\ for this reason be disentangled from the notion of rationality for which it cannot represent a plausible requirement.) \par
It is noteworthy that in the context of reinforcement learning (e.g.\ in robotics) different types of reward functions are usually formulated ranging from $ R(s') $ to $ R(s,a,s')$. For the purpose of ethical utility functions for advanced AI systems in critical application fields, we postulate that one does not have the choice to specify the abstraction level of the utility function, since for instance $ U(s') $ might lead to safety risks.  Christiano et al.\ ~\shortcite{christiano2017deep} considered the elicitation of human preferences on trajectory (state-action pairs) segments of a reinforcement learning agent i.a.\ realized by human feedback on short movies. For the purpose of utility elicitation in an AU framework exemplarily using a naive model as specified in equation (1), people will similarly have to assign utility to a movie representing a transition in the future (either in a mental mode or augmented by technology such as VR or AR~\cite{Delphi}). However, it is obvious that this naive utility assignment would not scale in practice. Moreover, it has not yet been specified how to aggregate ethical goal functions at a societal level. In the following Subsection~\ref{approx}, we will address these issues by proposing a practicable approximation of the utility function in (1) and a possible societal aggregation of this approximate solution.

 \subsection{Approximation, Aggregation and Validation}
 \label{approx}

 So far, it has been stated throughout the paper that one has to adequately increase the variety of a utility function meant to be ethical in order to avoid violations of human ethical intuitions and vulnerability to attackers crafting ethical adversarial examples against the model. However, it is important to note that despite the negatively formulated motivation of the approach, the aim is to craft a utility model $ U $ which represents a better model of human ethical intuitions in general, thus ranging from samples that are perceived as highly unethical to those that are assigned a high ethical desirability. In order to craft practical solutions that lead to optimal results, it might be advantageous to perform a thought experiment imagining a utopia and from that impose practical constraints on its viability. It might not seem realistic to deliberate a future \textit{utopia 1} as a sustainable society which is stable across a very large time interval in which every human being acts according to the ethical intuitions of all humans including the own and every artificial intelligent system fulfills the ethical intuitions of all humans. However, it seems more likely that within a \textit{utopia 2} being a stable society in which every human achieves a high level of a scientific definition of well-being (such as e.g.\ PERMA~\cite{seligman2012flourish}) with artificial agents acting as to maximize context-sensitive utility according to which (human or artificial) agents promoting the (measurable) well-being of human patients is regarded as the most utile type of events, the ethical intuitions of humans might tend to get closer to each other. The reason being that the variety of human moral judgements might interestingly \textit{decrease} since it is conceivable that they will tend to exhibit more similar prior experiences (all imprinted by well-being) and have more similar environments (full of stable people with a high level of well-being). The main factor drawing differences could be the body -- especially biological factors. However, the parameters related to interoception might be closer to each other, since all humans exhibit a high level of well-being which classically includes frequent positive affect. It is conceivable that with time, such a society could converge towards the utopia 1.\par In the following, we will denote the mentioned utopia-related ideal cognitive template of a (human or artificial) agent $ A $ performing an act $ w $ that contributes to the well-being of a human patient $ P $ with $ A\xrightarrow{\text{w}}P $ in analogy to the cognitive template of dyadic morality. (Thereby, $ A\xrightarrow{\text{w}}P $ is perceiver-dependent i.a.\ because psychological measures of well-being include subjective and self-reported elements such as e.g.\ life satisfaction or furthermore positive emotions~\cite{seligman2012flourish}.) Augmented utilitarianism foresees the need to at least depict a final goal at the abstraction level of a perceiver-dependent function on a transition as reflected in $ U_x(s,a,s') $. The ideal cognitive template $ A\xrightarrow{\text{w}}P $ formulated for utopia 2 by which it has been argued that a decrease in the variety of human morality might be achievable in the long-term exhibits an abstraction level that is compatible with $ U_x(s,a,s') $.\par
  A thinkable strategy for the design of a utility model $ U  $ that is robust against ethical adversarial examples and a model of human ethical intuitions is to try to adequately increase its variety using relevant scientific knowledge and to complementarily attempt to decrease the variety of human moral judgements for instance by considering $ A\xrightarrow{\text{w}}P $ as high-level final goal such that the described utopia 2 ideally becomes a self-fulfilling prophecy. For it to be realizable in practice, we suggest that the appropriateness of a given aggregated societal ethical goal function could be approximately validated against its quantifiable impact on well-being for society across the time dimension. Since it seems however unfeasible to directly map all important transitions of a domain to their effect on the well-being of human entities, we propose to consider perceiver-specific and domain-specific utility functions indicating combined preferences that each perceiver $ x $ considers to be relevant for well-being from the viewpoint of $ x $ himself in that specific domain. For these combined utility functions to be grounded in science, they will have to be based on scientifically measurable parameters. We postulate that a possible aggregation at a societal level could be performed by the following steps: 1) agreement on a common validation measure of an ethical goal function (for instance the temporal development of societal satisfaction with AI systems in a certain domain or with future AGI systems, their aptitude to contribute to sustainable well-being), 2) agreement on \textit{superset} $ O $ of scientifically measurable and relevant parameters (encoding e.g.\ affective, dyadic, cultural, social, political, socio-geographical but importantly also law-relevant information) that are considered as important across the whole society, 3) specification of personal utility functions for each member $ n $ of a society of $ N $ members allowing personalized and tailored combinations of a subset of $ O $, 4) aggregation to a societal ethical goal function $ U_{Total}(s,a,s') $. Taken together, these considerations lead us to the following possible approximation for an aggregated societal ethical goal function given a domain:

 \begin{align}
U_{Total}(s,a,s')= \sum_{n=1}^{N} \sum_{i=1}^{j} w_{i}^nf_{fi}^n(C_i)
\end{align}%
 with $ N $ standing for the number of participating entities in society, $ C_i= (p_{i1}, p_{i2},...,p_{im}) $ being a cluster of $ m\geq 1 $ correlated parameters (whereby independent factors are assigned an own cluster each) and $ f $ representing a set of preference functions (\textit{form functions}). For instance $ f=\{f_1,f_2,...,f_f\} $ where $ f_1 $ could be a linear transformation, $ f_2 $  a concave, $ f_3 $ a convex preference function and so on. Each entity $ n $ assigns a weight $ w_{i}^n  $ to a form function $ f_{fi}^n $ applied to a cluster of parameters $ C_i  $ whereby $\sum_{i=1}^{j} w_{i}^n=1 $. We define $ O=\{C_1, C_2,...\} $ as the superset of all parameters considered in the overall aggregated utility function. Moreover, $ a \in A $ with $ A $ representing the foreseen discrete action space at the disposal of the AI. (It is important to note that while the AI could directly perform actions in the environment, it could also be used for policy-making and provide plans for human agents.) Further, we consider a continuous state space with the states $ s$ and $s' \in S = \mathbb{R}^{|O|} $.  Other aspects including e.g.\ legal rules and norms on the action space can be imposed as constraints on the utility function. In a nutshell, the utility aggregation process can be understood as a voting process in which each participating individual $ n $ distributes his vote across scientifically measurable clusters of parameters $ C_i $ on which he applies a preference function $ f_{fi}^n $ to which weights $ w_{i}^n  $ are assigned as identified as relevant by $ n $ given a to be approximated high-level societal goal (such as $ A\xrightarrow{\text{w}}P $). In short, people do not have to agree on personal preferences and weightings, but only on a superset of acceptable parameters, an aggregation method and an overall validation measure. (Note that instead of involving society as a whole for each domain, the utility elicitation procedure can as well be approximated by a transdisciplinary set of representative experts (e.g.\ from the legislative) crafting \textit{expert ethical goal functions} that attempt to ideally emulate $ U_{Total}(s,a,s') $).\par Finally, it is important to note that the societal ethical goal function specified in (2) will need to be updated (and evalutated) at regular intervals due to the mental-state-dependency of utility entailing time-dependency~\cite{aliman2019augmented}. This leads to the necessity of a socio-technological feedback-loop which might concurrently offer the possibility of a \textit{dynamical ethical enhancement}~\cite{Delphi,werkhoven2018telling}. Pre-deployment, one could in the future attempt a validation via selected preemptive simulations~\cite{Delphi} in which (a representation of) society experiences simulations of future events $ (s,a,s')  $ as movies, immersive audio-stories or later in VR and AR environments. During these experiences, one could approximately measure the temporal profile of the so-called \textit{artificially simulated future instant utility}~\cite{Delphi} denoted  $ U_{TotalAS} $ being a potential constituent of future well-being. Thereby, $ U_{TotalAS} $ refers to the instant utility~\cite{kahneman1997back} experienced during a technology-aided simulation of a future event whereby instant utility refers to the affective dimension of valence at a certain time $ t $. The temporal integral that a measure of $ U_{TotalAS} $ could approximate is specified as:
 \begin{align}
 U_{TotalAS}(s,a,s')\approx  \sum_{n=1}^{N}\int_{t_0}^{T} I_n(t) dt
 \end{align}%
 with $ t_0  $referring to the starting point of experiencing the simulation of the event $ (s,a,s') $ augmented by technology (movie, audio-story, AR, VR) and $ T $ the end of this experience.
$ I_n(t) $ represents the valence dimension of core affect experienced by $ n $ at time $ t $. Finally, post-deployment, the ethical goal function of an AI system can be validated using the validation measure agreed upon before utility aggregation (such as the temporal development of societal-level satisfaction with an AI system, well-being or even the perception of dyadicness) that has to be a priori determined.

\section{Conclusion and Future Work}
\label{conc}
In this paper, we motivated the need in AI value alignment to attempt to model utility functions capturing the variety of human moral judgements through the integration of relevant scientific knowledge -- especially from neuroscience and psychology -- (instead of learning) in order to avoid violations of human ethical intuitions. We reformulated value alignment as a security task and introduced the requirement to increase the variety within classical utility functions positing that a utility function which does not integrate affective and perceiver-dependent dyadic information does not exhibit sufficient variety and might not exhibit robustness against corresponding adversaries. Using augmented utilitarianism as a suitable non-normative ethical framework, we proposed a methodology to implement and possibly validate societal perceiver-dependent ethical goal functions with the goal to better incorporate the requisite variety for AI value alignment.\par In future work, one could extend and refine the discussed methodology, study a more systematic validation approach for ethical goal functions and perform first experimental studies. Moreover, the \textit{``security of the utility function itself is essential, due to the possibility of its modification by malevolent actors during the deployment phase"}~\cite{aliman2019augmented}. For this purpose, a blockchain-based solution might be advantageous. In addition, it is important to note that even with utility functions exhibiting a sufficient variety for AI value alignment, it might still be possible for a malicious attacker to craft adversarial examples against a utility maximizer at the perception-level which might lead to unethical behavior. Besides that, one might first need to perform policy-by-simulation~\cite{werkhoven2018telling} prior to the deployment of advanced AI systems equipped with ethical goal functions for safety reasons. Last but not least, the usage of ethical goal functions might represent an interesting approach to the AI coordination subtask in AI Safety, since an international use of this method might contribute to reduce the AI race to the problem-solving ability dimension~\cite{Delphi}.

\bibliographystyle{named}
\bibliography{ijcai19}

\end{document}